%% file: melba_focused_decoder.tex
\documentclass[twoside,11pt]{article}

\usepackage[accepted]{melba}

\usepackage{mwe} % to get dummy images, only for the example
\usepackage{amsmath,amsfonts}
\usepackage{url}
\usepackage{multirow}
\usepackage{pifont}
\usepackage{float}
\usepackage{booktabs}
\usepackage{makecell}

\melbaid{2023:003}  % This is provided upon by the publishing editor
\melbaauthors{Wittmann, Navarro, Shit, and Menze}
\volume{2}
\firstpageno{72}  % Communicated by the publishing editor
\year{2023}  % The publication year
\datesubmitted{10/2022}  % Date submitted to MELBA: mm/yyyy
\datepublished{02/2023}  % Today's date: mm/yyyy\

\ShortHeadings{Focused Decoding Enables 3D Anatomical Detection by Transformers}{Wittmann, Navarro, Shit, and Menze}

\title{Focused Decoding Enables \\ 3D Anatomical Detection by Transformers}

\author{\name Bastian Wittmann \email bastian.wittmann@uzh.ch \\  % start right after \author{, or there will be an extra space
	\addr Department of Quantitative Biomedicine, University of Zurich, Zurich, Switzerland
	\AND
	\name Fernando Navarro \email fernando.navarro@tum.de \\
	\addr Department of Informatics, Technical University of Munich, Munich, Germany
    \AND
	\name Suprosanna Shit \email suprosanna.shit@tum.de \\
	\addr Department of Informatics, Technical University of Munich, Munich, Germany
    \AND
	\name Bjoern Menze \email bjoern.menze@uzh.ch \\
	\addr Department of Quantitative Biomedicine, University of Zurich, Zurich, Switzerland
}

\begin{document}

% top matter
\maketitle

% abstract
\begin{abstract}%   <- trailing '%' for backward compatibility of .sty file
\input{sections/00_abstract}
Our code is available at~\url{https://github.com/bwittmann/transoar}.
\end{abstract}

% keywords
\begin{keywords}
    Anatomical Structure Localization, Attention, CT, Deep Learning, Detection Transformer, Explainable AI
\end{keywords}

\section{Introduction}
\input{sections/01_intro.tex}\label{sec:introduction}

\section{Related Works}
\input{sections/02_related_lit.tex}\label{sec:related_work}

\section{Methods}\label{sec:method}
\input{sections/03_method.tex}

\section{Experiments and Results}\label{sec:experiments}
\input{sections/04_exp_res.tex}

\section{Limitations}\label{sec:limitations}
\input{sections/05_limitations.tex}

\section{Outlook and Conclusion}\label{sec:outlook}
\input{sections/06_outlook.tex}
\acks{This work has been supported in part by the TRABIT under the EU Marie Sklodowska-Curie Program (Grant agreement ID: 765148) and in part by the DCoMEX (Grant agreement ID: 956201). The work of Bjoern Menze was supported by a Helmut-Horten-Foundation Professorship.}

% Ethical Standards.
% Please edit with the appropriate ethics considerations for your work. Include any pertinent IRB information, etc.
%
% Please note that the submission requirements included:
% The work presented must follow appropriate ethical standards in conducting research and writing the manuscript, following all applicable laws and regulations regarding treatment of animals or human subjects.
\ethics{The work follows appropriate ethical standards in conducting research and writing the manuscript, following all applicable laws and regulations regarding treatment of animals or human subjects.}

% Conflict of Interest
% Declaration of possible conflicts of interest: Authors must disclose any financial, organisational, commercial or personal conflicts of interest that might bias their work.
% If no conflicts, please say "We declare we don't have conflicts of interest."
\coi{We declare we don't have conflicts of interest.}

\bibliography{melba_focused_decoder}

% Manual newpage inserted to improve layout of sample file - not
% needed in general before appendices.
\newpage
\end{document}

%% file: sections/00_abstract.tex
Detection Transformers represent end-to-end object detection approaches based on a Transformer encoder-decoder architecture, exploiting the attention mechanism for global relation modeling. Although Detection Transformers deliver results on par with or even superior to their highly optimized CNN-based counterparts operating on 2D natural images, their success is closely coupled to access to a vast amount of training data. This, however, restricts the feasibility of employing Detection Transformers in the medical domain, as access to annotated data is typically limited. To tackle this issue and facilitate the advent of medical Detection Transformers, we propose a novel Detection Transformer for 3D anatomical structure detection, dubbed \emph{Focused Decoder}. \emph{Focused Decoder} leverages information from an anatomical region atlas to simultaneously deploy \emph{query anchors} and restrict the cross-attention's field of view to regions of interest, which allows for a precise focus on relevant anatomical structures. We evaluate our proposed approach on two publicly available CT datasets and demonstrate that \emph{Focused Decoder} not only provides strong detection results and thus alleviates the need for a vast amount of annotated data but also exhibits exceptional and highly intuitive explainability of results via attention weights.

%% file: sections/01_intro.tex
% why detection
Over time, object detection has firmed its position as a fundamental task in computer vision~\citep{coco, adek, kitti}. In medical imaging, however, object detection remains heavily under-explored, as the primary focus lies on the dominant discipline of semantic segmentation. This can be mostly traced back to the fact that many clinically relevant tasks require voxel-wise predictions. Object detection, on the other hand, yields object-specific bounding boxes as output.
Even though these bounding boxes themselves are of relevancy for object-level diagnostic decision-making~\citep{nndet}, efficient medical image retrieval and sorting~\citep{criminisi2010regression}, streamlining complex medical workflows~\citep{gauriau2015multi}, and robust quantification~\citep{tong2019disease}, they can be additionally utilized to increase the performance of many medically relevant downstream tasks, such as semantic segmentation~\citep{det_framework, liang2019deep}, image registration~\citep{samarakoon2017light}, or lesions detection~\citep{mamani2017organ}. In this context, the bounding boxes generated by the preliminary detection step allow the downstream task to focus solely on regions that are likely to contain relevant information, which not only introduces a strong inductive bias but also improves the computation efficiency and unifies the sizes of regions for further analysis.
In the specific case of semantic segmentation, the estimated bounding boxes are utilized to crop organ-wise RoIs from the original CT images. Subsequently, these organ-wise RoIs are used to train organ-specific segmentation networks allowing the individual segmentation networks to solely learn highly specialized organ-specific feature representations. This results in a binary instead of a multi-class semantic segmentation problem.
However, since this hypothesis only holds if the detection algorithm yields precise results, special attention must be paid towards the development of medical detection algorithms.

% This applies in particular to the segmentation of regular anatomical structures rather than anomalies such as lesions. While lesions are generally represented in CT scans by drastic local intensity changes, intensities of internal structures often lie in close proximity. Hence information encoded in structure boundaries and co-occurrence is of utmost importance for the segmentation of internal structures. Since most segmentation algorithms, however, rely on a sliding window inference approach to process CT images of high resolution, they fail to capture this valuable information as their voxel-wise predictions are mostly driven by local intensity changes and texture. Therefore, we argue that forwarding object-specific bounding boxes to the segmentation algorithm, which processes these received regions of interest in parallel, mitigates these issues, as we exploit information implicitly encoded in the bounding boxes, reduce the region to analyze, and lower the complexity from a multi-class semantic segmentation problem to a binary segmentation problem. However, since this hypothesis only holds if the detection algorithm yields precise results, special attention must be paid towards the development of 3D medical object detection algorithms.\\
% 2D natural image detection advances
Currently, most state-of-the-art architectures operating on 2D natural images exploit the global relation modeling capability of Vision Transformers. These Vision Transformers can generally be divided into two categories, namely Vision Transformer Backbones and Detection Transformers. While Detection Transformers represent complete end-to-end object detection pipelines with an encoder-decoder structure similar to the original machine translation Transformer~\citep{attn_all}, Vision Transformer Backbones solely utilize the Transformer's encoder for feature refinement. To obtain competitive results, object detectors typically combine variants of Vision Transformer Backbones with two-stage~\citep{faster_rcnn} or multi-stage approaches~\citep{htc} based on convolutional neural networks (CNN).

% To obtain competitive results, object detectors typically combine variants of Vision Transformer Backbones with two-stage approaches like Faster R-CNN~\cite{faster_rcnn}, which leverages region-proposal and RoI-pooling layers to extract intermediate representations, or advanced multi-stage approaches like HTC++~\cite{htc, swinv2}.
%Additionally, augmenting these traditional approaches with advanced classification and bounding box regression head networks~\cite{dyn_heads} has shown to further increase detection performance.\\ 
Although the success of Vision Transformer Backbones has shown that the global relation modeling capability introduced by the attention operation is clearly beneficial, the novel concept of Detection Transformers was unable to produce competitive performances and, therefore, mostly seen as an alternate take on object detection, particularly suitable for set prediction. However, as time evolved, modifications~\citep{def_detr,dn_detr,dab_detr} of the original Detection Transformer~\citep{detr} narrowed the performance gap to their highly optimized and intensively researched CNN-based counterparts until a Detection Transformer, called DINO~\citep{dino}, was finally able to achieve state-of-the-art results on the COCO benchmark.
% Although Vision Transformer Backbones quickly caught up to the convolutional counterpart, thanks to the success of global relational modeling, Detection Transformers remained for some time as an alternate take on object detection particularly suitable for set prediction because of its performance gap from the CNN-based model. However, as time evolved, modifications \cite{def_detr,dn_detr,dab_detr} of the original Detection Transformer~\cite{detr} narrowed down the performance gap to traditional CNN-based detectors until a Detection Transformer variant, called DINO~\cite{dino}, was finally able to achieve state-of-the-art results on the COCO benchmark.

% current 3D medical detection
Despite the successful employment of Detection Transformers for computer vision tasks processing 2D natural images, the feasibility of adapting Detection Transformers for medical detection tasks remains largely unexplored. The faltering progress of Detection Transformers in the medical domain can be mainly attributed to the absence of large-scale annotated datasets, which are crucial for the successful employment of Detection Transformers~\citep{vit}. This is because their attention modules require long training schedules to learn sparse, focused attention weights due to their shortage of inductive biases compared to the convolutional operation. In this regard, large-scale datasets are essential to avoid overfitting.
Therefore, current medical object detectors still rely predominantly on CNN-based approaches.

% \hl{
% This is because their attention modules require long training schedules to learn sparse attention weights, which are required to focus on relevant information. Therefore, to avoid overfitting, large-scale datasets are essential. The slow convergence of Detection Transformers can be attributed to their shortage of inductive biases compared to the convolutional operation. The fact that medical object detection is usually formulated as a 3D detection problem additionally stresses the Detection Transformers lack of inductive.
% }.
% Therefore, current medical object detectors still rely predominantly on CNN-based approaches.
% % combined with the fact that medical object detection is usually formulated as a 3D detection problem

% why use detection transformers for 3D medical detection
In order to accelerate the advent of Detection Transformers to the medical domain, systematic research focusing on their adaption and thus overcoming the limitations imposed by small-scale datasets has to be conducted.
Many clinically relevant detection tasks such as organs-at-risk or vertebrae detection rely on a well-defined anatomical field of view (FoV)~\citep{schoppe2020deep}. This implies that approximate relative and absolute positions of anatomical structures of interest remain consistent throughout the whole dataset. For example, the right lung always appears above the liver in the top right region. 
We argue that Detection Transformers should bear an immense performance potential for these well-defined FoV anatomical structure detection tasks by presenting three major arguments:
% We present three major arguments below in favour of the use of medical Detection Transformers for well-defined FoV detection tasks, besides the immense potential demonstrated by Detection Transformer on 2D natural images. 
% \emph{First}, the Detection Transformer's learned positional embeddings, also called object queries, should be especially beneficial in the context of well-defined FoVs;
\begin{enumerate}
\item The Detection Transformers' learned positional embeddings, also called query embeddings, should be especially beneficial in the context of the positional consistency assumption of well-defined FoV detection tasks.\label{mot:first}

\item The concept of relation modeling inherent to Detection Transformers should allow the model to capture relative positional inter-dependencies among anatomical structures via self-attention among object queries.\label{mot:second}

\item Visualization of attention weights allows for exceptional and accessible explainability of results, which is a crucial aspect of explainable artificial intelligence in medicine. \label{mot:third}
\end{enumerate}

% what is done in this work
Therefore, this work aims to pave the path for Detection Transformers, solving the task of 3D anatomical structure detection. We propose a novel Detection Transformer, dubbed \emph{Focused Decoder}, that alleviates the need for large-scale annotated datasets by reducing the complexity of the 3D detection task. To this end, \emph{Focused Decoder} builds upon the positional consistency assumption of well-defined FoV detection tasks by exploiting anatomical constraints provided by an anatomical region atlas. \emph{Focused Decoder} utilizes these anatomical constraints to restrict the cross-attention’s FoV and reintroduce the well-known concept of anchor boxes in the form of \emph{query anchors}. We prove that \emph{Focused Decoder} drastically outperforms two Detection Transformer baselines~\citep{detr,def_detr} and performs comparably to the CNN-based detector RetinaNet~\citep{retinaunet} while having a significantly lower amount of trainable parameters and increased explainability of results. 
We summarize our contribution as follows:
\begin{itemize}
 \item To the best of our knowledge, we are the first to successfully leverage 3D Detection Transformers for 3D anatomical structure detection.
 
 \item We adapt two prominent 2D Detection Transformers, DETR and Deformable DETR, for 3D anatomical structure detection.
 
 \item We introduce \emph{Focused Decoder}, a novel Detection Transformer producing strong detection results for well-defined FoV detection tasks by exploiting anatomical constraints provided by an anatomical region atlas to deploy \emph{query anchors} and restrict the cross-attention’s FoV.
 
  \item We identify attention weights as a crucial aspect of explainable artificial intelligence in medicine and demonstrate that \emph{Focused Decoder} improves the explainability of results.
 
 \item We compare performances of DETR, Deformable DETR, and \emph{Focused Decoder} with CNN-based approaches represented by RetinaNet on two publicly available computed tomography (CT) datasets and ensure direct comparability of results.
\end{itemize}

% Supros comment regarding lesion vs reg. structure:
%The point you made here is fantastic. You can also expand on the regular anatomy point that you brought up in the last meeting: For example: There two main kind of image segmentation, regular anatomy segmentation like yours and anomaly segmentation like brain lesion, while the anomaly mostly based on local contrast change it is easier to go in the segmentation direction and object detection is an over kill, however when someone wants to segment regular structure, intensities in CT image for different organs lie in close proximity, hence instead of local contrast and texture, object boundary and co-occurrence of different object is much more important. (you can cite Anjany's VERSE paper here to make the point that for large scale regular structure direct segmentation fails, whereas detection-segmentation pipeline is the winner ) hence object detection is not only helping segmentation but also unavoidable to some extent

% this is a bit sudden, you have to give some intro to reviewer from where it is coming from: maybe something like: We experiment with DETR and def-DETr and identified the scope and importance of exclusively focusing decoder's attention to organ specific regions. To this end, we porpose a novel architecture ........

%% file: sections/02_related_lit.tex
Since \emph{Focused Decoder} builds upon prior work, we discuss related work on 2D Detection Transformers processing natural images followed by 3D medical detection algorithms. We pay special attention to detectors featured in this work.

\subsection{2D Detection Transformers}
DETR~\citep{detr} laid the foundations for Detection Transformers. DETR adopts the machine translation Transformer's~\citep{attn_all} encoder-decoder architecture to create a streamlined end-to-end object detection pipeline. However, in contrast to the machine translation Transformer, DETR predicts the final set of bounding boxes in parallel rather than in an autoregressive manner. DETR employs a ResNet-based backbone to generate a low resolution representation of the input image, which is subsequently flattened and refined in $N$ Transformer encoder blocks. The refined input sequence is in the next step forwarded to the decoder's cross-attention module to refine a set of object queries. After the object queries have been refined in $N$ Transformer decoder blocks, they are fed into a classification and a bounding box regression head, resulting in predicted bounding boxes and class scores. During training, a Hungarian algorithm enforces one-to-one matches, which are particularly suitable for set prediction tasks, based on a set of weighted criteria. 
% This implies that a small subset of object queries is matched to the present ground truth objects, while the rest is assigned to the background class.
Although DETR's simple design achieves promising 2D detection results on par with a Faster R-CNN baseline, it suffers from high computational complexity, low performance on small objects, and slow convergence.

Deformable DETR~\citep{def_detr} improved DETR's detection performance while simultaneously reducing the model's computational complexity and training time by introducing the deformable attention mechanism. The concept of deformable attention has been derived from the concept of deformable convolution~\citep{def_conv}, which increases the modeling ability of CNNs by leveraging learned sampling offsets from the original grid sampling locations. The deformable attention module utilizes the learned sparse offset sampling strategy introduced by deformable convolutions by allowing an object query to solely attend to a small fixed set of sampled key points and combines it with the relation modeling ability of the attention operation. 
% Besides the use of the deformable attention mechanism, Deformable DETR is also slightly modified compared to DETR. In particular, Deformable DETR predicts relative offsets to previously predicted reference points.

It is worth mentioning that many Detection Transformer variants tried to improve DETR's initial concept over time. For example, Efficient DETR~\citep{efficient_detr} eliminated DETR's need for an iterative object query refinement process, Conditional DETR~\citep{conditional_detr} introduced a conditional cross-attention module, DAB-DETR~\citep{dab_detr} updated the object query formulation to represent stronger spatial priors, DN-DETR~\citep{dn_detr} presented a denoising training strategy, and DINO~\citep{dino} combined and improved important aspects like denoising training, query initialization, and box prediction.

\subsection{3D Medical Detection}
Even though 3D medical detection is a longstanding topic in medical image analysis~\citep{criminisi2009decision, criminisi2010regression}, most research currently focuses on the more dominant discipline of semantic segmentation~\citep{isensee2021nnu, navarro2021evaluating, navarro2019shape}. This is because most clinically relevant tasks require voxel-vise predictions. Therefore, prior work on deep learning-based 3D medical detection remains relatively limited.

\subsubsection{CNN-Based Detectors}
Apart from individual experiments with different detectors, such as 3D Faster R-CNN~\citep{medical_faster_rcnn}, most research focused on the CNN-based detector Retina U-Net~\citep{retinaunet, nndet}. Retina U-Net builds upon the one-stage detector RetinaNet~\citep{retinanet} and modifies its architecture for 3D medical detection. Retina U-Net's main contribution is represented by the introduction of a segmentation proxy task. The segmentation proxy task allows the use of voxel-level annotations present in most medical datasets as an additional highly detailed supervisory signal. For detecting objects at different scales, Retina U-Net's backbone, given by a modified feature pyramid network (FPN), forwards multi-level feature maps to the bounding box regression and classification head networks. Since each voxel of these multi-level feature maps is assigned to a set of 27 level-specific anchor boxes varying in scale, the bounding box regression head predicts position and size offsets, while the classification head predicts their corresponding class scores. As Retina U-Net, therefore, represents a dense one-stage detection scheme, it utilizes 
% adaptive training sample selection~\citep{atss} for matching during training and
non-maximum suppression to reduce duplicates during inference. Retina U-Net is also featured in the automated medical object detection pipeline nnDetection~\citep{nndet}. Following nnU-Net's~\citep{isensee2021nnu} agenda, nnDetection adapts itself without any manual intervention to arbitrary medical detection problems while achieving results on par with or superior to the state-of-the-art. Furthermore, SwinFPN~\citep{swinfpn} experimented with Vision Transformer Backbones by incorporating 3D Swin Transformer blocks~\citep{video_swin} in Retina U-Net's architecture, following recent trends in medical semantic segmentation~\citep{unetr, swin_unetr}. %, cotr, nnformer, vt_unet}.
% Xu~\emph{et al.} proposed a 3D Faster R-CNN framework~\cite{medical_faster_rcnn} exploiting a 3D region proposal network for multiple organ detection. Their approach is fully implemented in a 3D manner and utilizes a novel feature extraction backbone generating rich high-resolution feature maps.\\
% In 3D medical imaging, Vision Transformer Backbones were mostly utilized for the task of semantic segmentation~\cite{unetr, swin_unetr, cotr, nnformer, vt_unet}. Following this trend, Wittmann~\emph{et al.} proposed SwinFPN~\cite{swinfpn}, a lightweight FPN incorporating 3D Swin Transformer blocks~\cite{swin, video_swin} in the down-sampling branch. SwinFPN's authors exchange Retina U-Nets standard FPN backbone with SwinFPN and validate its performance for the task of organs-at-risk detection. They show that SwinFPN combined with the CNN-based detector Retina U-Net results in increased detection performances.

\subsubsection{Detection Transformers}
Although some studies experimented with Detection Transformers operating on 2D medical data~\citep{cell_detr, cotr_detr}, only a few attempts tried to adapt them for 3D medical detection tasks. Spine-Transformer~\citep{spinetr}, for example, leverages DETR for sphere-based vertebrae detection. To this end, Spine-Transformer augments DETR with skip connections and additional learnable positional encodings. In contrast to \emph{Focused Decoder}, however, Spine-Transformer's concept relies on data of arbitrary FoV.
%on two publicly available datasets~\cite{spinetr_dataset1, verse}.
Relationformer~\citep{relationformer} successfully utilizes Deformable DETR to detect small-scale blood vessels from 3D voxel-level segmentations. The authors introduce an additional relation-token and demonstrate that Relationformer achieves state-of-the-art performances for multi-domain image-to-graph generation tasks, such as blood vessel graph generation.
% The authors introduce a novel relation token, learning object interaction in the context of semantic or global reasoning.\\
% Recently, Liu~\emph{et al.}\cite{dual_modality_detr} tried to tackle the 3D multiple organ detection task by training a 2D Detection Transformer similar to DETR on 2D projections of volumetric CT and PET scans. To generate the final 3D bounding boxes, the predicted, multi-view 2D bounding boxes are subsequently back-projected. Additionally, the employed Detection Transformer assigns to each target organ an individual object query, which simplifies the task and facilitates faster convergence.

%% file: sections/03_method.tex
This section builds our rationale for the proposed Detection Transformer \emph{Focused Decoder}.
% , refining a set of object queries for detecting anatomical structures.
\emph{Focused Decoder} explicitly takes advantage of the fact that approximate relative and absolute positions of labeled anatomical structures contained in datasets of well-defined FoV are consistent, which we try to ensure in an abundant and necessary preprocessing step. We, therefore, draw parallels to the well-known concept of anatomical atlases~\citep{hohne1992volume}. Based on this positional consistency assumption, we first determine for each dataset FoV-specific anatomical region atlases containing regions of interest (RoI) that comprise labeled anatomical structures. Subsequently, we place in each structure- or class-specific RoI uniformly spaced \emph{query anchors} and assign a dedicated object query to each of them, resulting in two levels of abstraction. Therefore, \emph{Focused Decoder's} object queries are not only assigned to individual RoIs but also to \emph{query anchors} located inside  RoIs. This allows \emph{Focused Decoder} to restrict the object queries' FoV for cross-attention to solely voxel within their respective RoI, simplify matching during training and inference, and overcome patient-to-patient variability by generating diverse class-specific predictions enforced by \emph{query anchors}.

In the following, we introduce the FPN feature extraction backbone and the atlas generation process. Finally, we present the novel concept of \emph{query anchors} and essential aspects of \emph{Focused Decoder}, including its architecture, the \emph{focused cross-attention module}, the concept of relative offset prediction, its matching strategy, and its loss function.

\subsection{FPN Feature Extraction Backbone}
\begin{figure}[t!]
\centerline{\includegraphics[width=0.50\columnwidth]{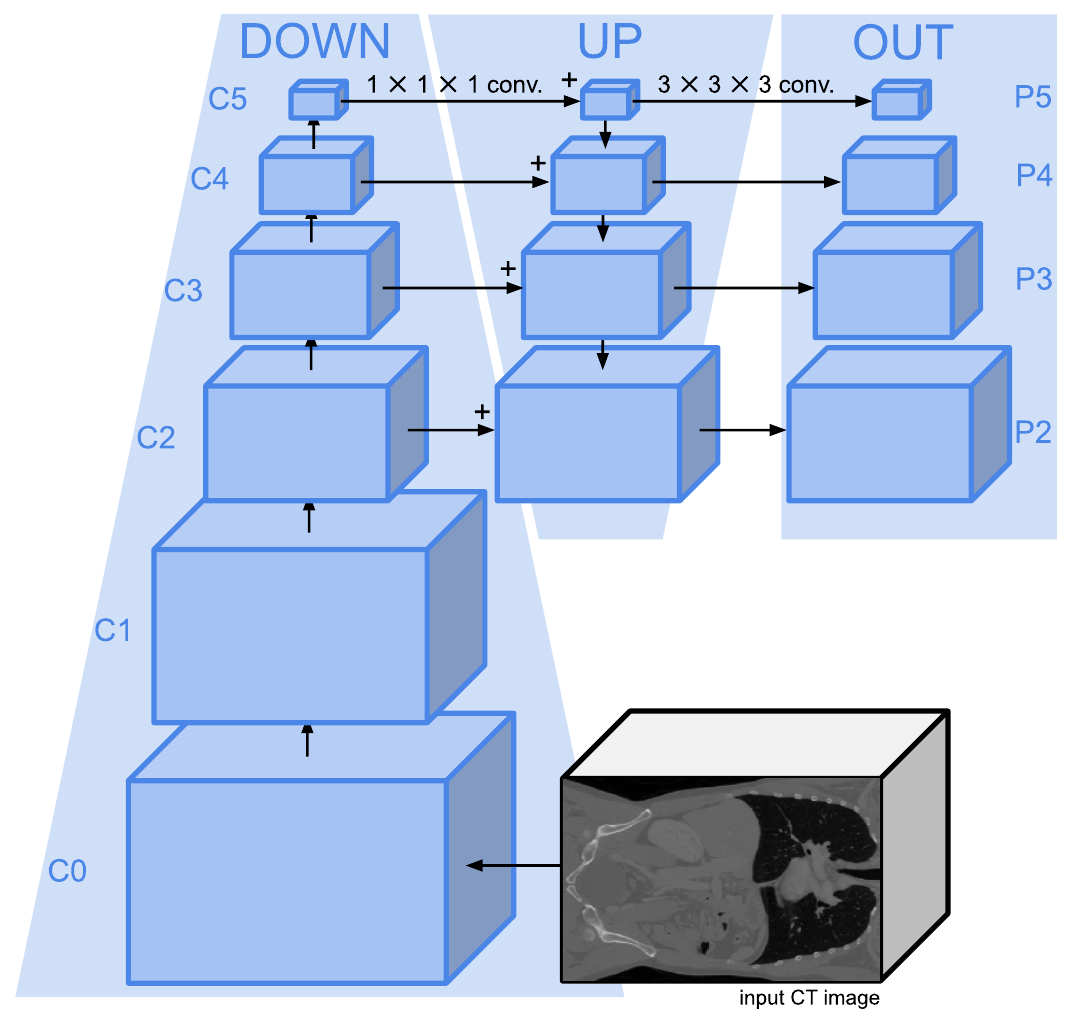}}
\vspace{-1.0em}
\caption{Detailed structure of the FPN. The FPN can generally be divided into three sections: the down-sampling branch (down, C0 - C5), the up-sampling branch (up), and the final output projection (out, P2 - P5).}
\vspace{-1.5em}
\label{fig:fpn}
\end{figure}
Following DETR, \emph{Focused Decoder} relies on a feature extraction backbone to create a lower resolution representation of the input CT image $x \in\mathbb{R}^{H \times W \times D}$, which mitigates the high computational complexity of the attention operation. This feature extraction backbone is depicted in Fig.~\ref{fig:fpn} and is given, similar to Retina U-Net, by an FPN. It should be mentioned that although the feature extraction backbone serves as a CNN-based feature encoder, we refrain from referring to it as an encoder to avoid confusion between the Transformer's encoder and the FPN.
The FPN consists of a CNN-based down-sampling branch (down), an up-sampling branch (up), and a final output projection (out). The up-sampling branch incorporates the down-sampled multi-level feature maps $\{ x_{Cl} \}_{l=2}^{5}$, where $x_{Cl} \in\mathbb{R}^{24 \cdot 2^{l}\times H/2^{l} \times W/2^{l} \times D/2^{l}}$, via lateral connections based on $1 \times 1  \times 1$ convolutions and combines them with up-sampled feature maps of earlier stages. A final output projection acts as an additional refinement stage, fixing the channel dimension to $d_{\text{hidden}}$. Therefore, the FPN outputs a set of refined multi-level feature maps $\{x_{Pl} \}_{l=2}^{5}$, where $x_{Pl} \in\mathbb{R}^{d_{\text{hidden}}\times H/2^{l} \times W/2^{l} \times D/2^{l}}$, encoding semantically strong information by leveraging the top-down inverted high- to low-level information flow. Even though \emph{Focused Decoder} solely processes the feature map $x_{P2}$, we require multi-level feature maps for our experiments. This is because our experiments rely on the same feature extraction backbone to ensure maximum comparability of results (see Fig.~\ref{fig:method_overview} and Table~\ref{tab:ablations_fmap}).
% The standard down-sampling block in the down-sampling branch consists of two consecutive $3 \times 3 \times 3$ convolutional layers, which are each followed by instance normalization and a ReLU non-linearity, while the up-sampling block in the top-down branch is given by a $3 \times 3 \times 3$ transposed convolutional layer.

\begin{figure}[t]
\centerline{\includegraphics[width=0.65\columnwidth]{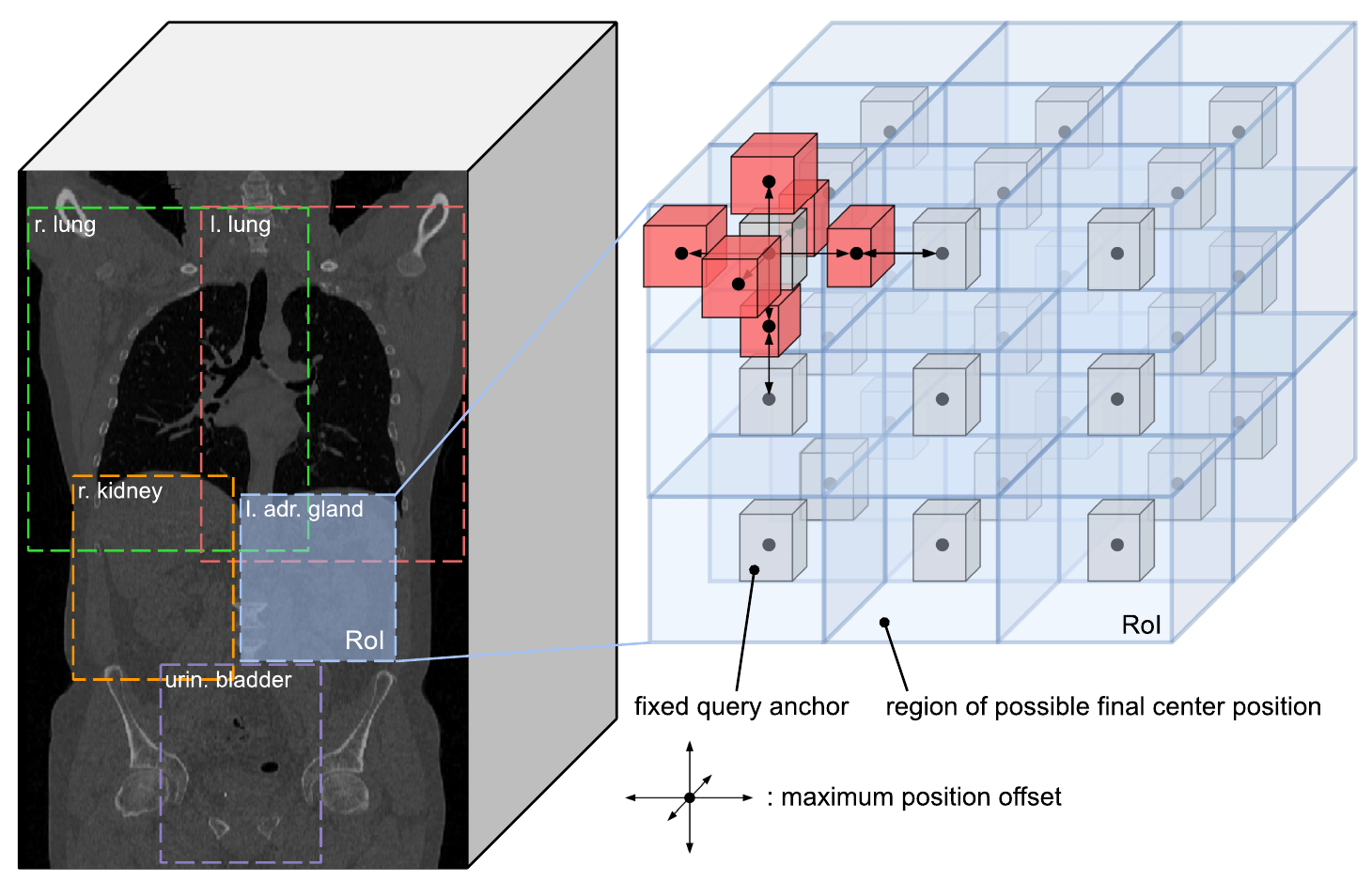}}
\vspace{-1.0em}
\caption{Concept of \emph{query anchors}. We place 27 \emph{query anchors} of class-specific median size at uniformly spaced positions in each RoI. \emph{Focused Decoder} predicts relative position and size offsets with regard to fixed \emph{query anchors}. Exemplary predictions corresponding to the upper left \emph{query anchor} of the left adrenal gland pushing the position offset restriction to its limit are visualized in red. Regions of possible query-specific final center positions are indicated in blue.}
% Although the center positions and sizes of query anchors are fixed, we object queries can modify their respective query anchors via the concept of relative offset prediction. Position offset restrictions of individual query anchors and maximum position offsets of a individual query anchor are indicated the in red.}
\label{fig:anchor_gen}
\vspace{-1.5em}
\end{figure}

\subsection{Atlas Generation}
In this paper, we generate custom atlases and refer to them as anatomical region atlases. Relying on the positional consistency assumption, we determine class-specific RoIs, describing the minimum volumes in which all instances of a particular anatomical structure are located. A small subset of RoIs is shown in Fig.~\ref{fig:anchor_gen} (left). Additionally, we estimate for each labeled anatomical structure the median, minimum, and maximum bounding box size, which will be necessary for the \emph{query anchor} generation process and the concept of relative offset prediction. Given that the test set should only be used to estimate the final performance of the model, we estimate RoIs and bounding box sizes solely based on instances contained in the training and validation sets.
% Although it would be optimal to rely on a drastically increased number of instances,  

% Although we try to ensure datasets consisting of images with a similar FoV in an abundant preprocessing step, this is often due to dataset-specific inconsistencies and the substantial amount of patient-to-patient variability with regard to position and sizes of anatomical structures difficult to achieve. This is reflected in RoIs, which occupy in comparison to the structures of interest larger regions. We argue that , allowing for enough flexibility to consistent capture the anatomic structure of interest, even for patients in the test set.

Since medical CT datasets typically contain different labeled anatomical structures, FoVs are inconsistent across datasets. Therefore, we generate dataset-specific anatomical region atlases. In general, however, existing anatomical atlases could be adjusted to fit the datasets at use based on a few anatomical landmarks~\citep{potesil2013learning, xu2016evaluation}. 
% Although we generate dataset-specific anatomical atlases in the context of this work, we would like to highlight that in theory existing anatomical atlases could be adjusted to fit the datasets at use based on anatomical landmarks. 

\subsection{Query Anchor Generation}
Due to dataset-specific inconsistencies and the substantial amount of patient-to-patient variability with regard to positions and sizes of anatomical structures, it is often difficult to ensure that datasets consist of images of exactly the same FoV. This is reflected in the fact that, on average, RoIs occupy 25 times larger volumes than anatomical structures. Therefore, \emph{Focused Decoder} further subdivides RoIs by assigning to each class-specific RoI in our anatomical region atlas 27 fixed class-specific \emph{query anchors} in the format $(c_x, c_y, c_z, h, w, d)$. We denote \emph{query anchors} by $q_{\text{anchors}} \in\mathbb{R}^{\#\text{classes} \times 27 \times 6}$. While the spatial locations of \emph{query anchors} are defined by uniformly spaced positions in their corresponding RoIs, their sizes are governed by the class-specific median bounding box sizes contained in the anatomical region atlas. Generated \emph{query anchors} associated with the left adrenal gland's RoI are shown in detail in Fig.~\ref{fig:anchor_gen} (right).
% These query anchors are in a later stage matched to object queries, which demystifies these learned positional embeddings by assigning them to precise spatial locations. In the context of this work, we utilize the generated query anchors for the sake of matching predictions to ground truth objects and to enable the concept of relative offset prediction.

\begin{figure}[ht]
\centerline{\includegraphics[width=1\columnwidth]{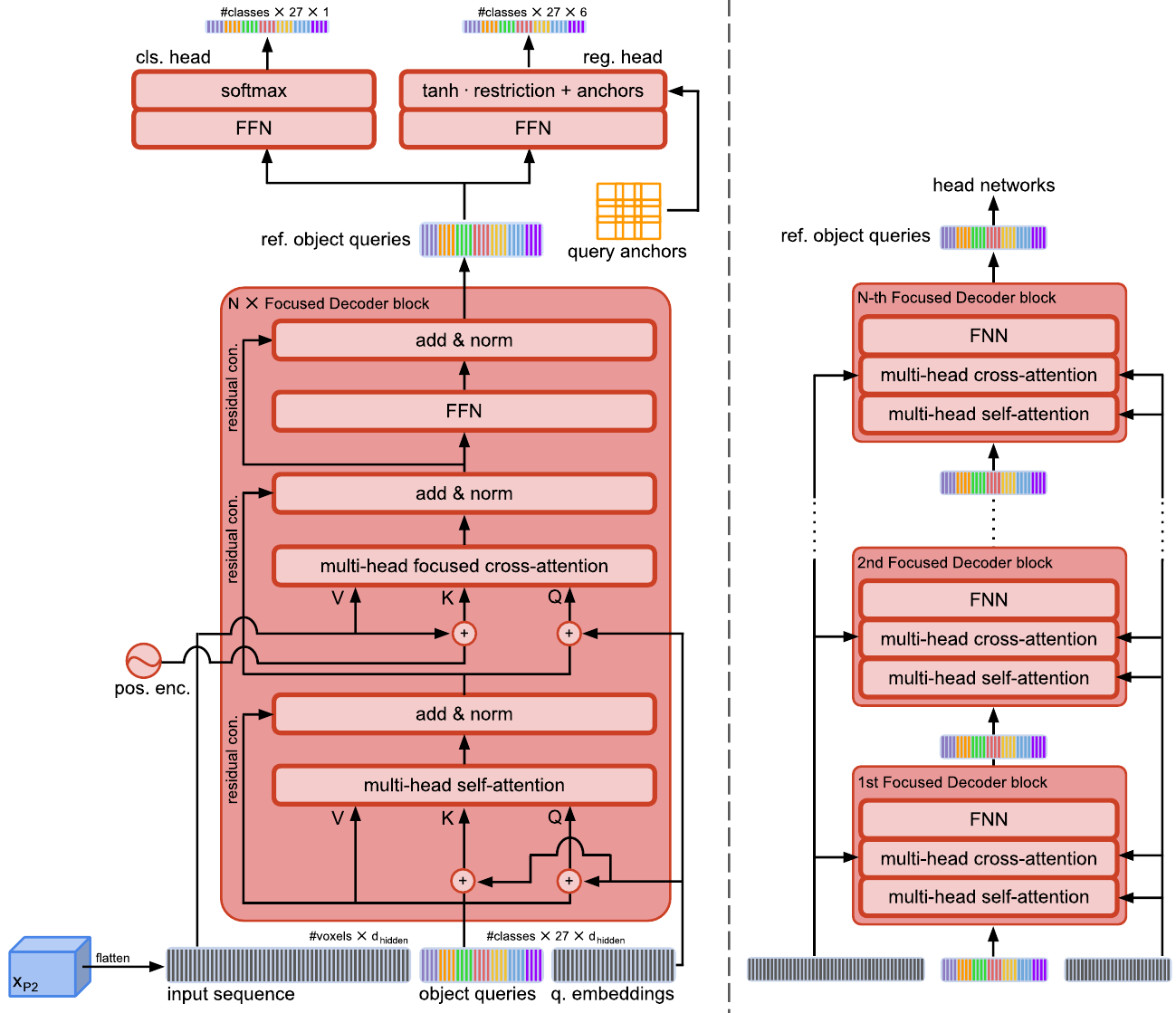}}
\caption{
(\textbf{Left}): \emph{Focused Decoder's} detailed architecture. The input sequence is given by the flattened feature map of the FPN stage P2. \emph{Focused Decoder} makes use of class-specific object queries by assigning object queries to 27 \emph{query anchors} located inside each class-specific RoI (see color coding).
Object queries are not only associated with \emph{query anchors} but also with learned positional embeddings, called query embeddings.
(\textbf{Right}): Iterative refinement process in $N$ stacked \emph{Focused Decoder} blocks. Object queries are initialized with zeros. Refined object queries are forwarded to subsequent \emph{Focused Decoder} blocks.
% The initial input to the first \mbox{\emph{Focused Decoder}} block is represented by zeros. For the following \mbox{\emph{Focused Decoder}} blocks, the inputs are given by the outputs of previous blocks.
% Therefore, \mbox{\emph{Focused Decoder's}} first self-attention module can be skipped.
}
\label{fig:foc_dec_arch}
\vspace{-1.5em}
\end{figure}

\subsection{Focused Decoder}
\subsubsection{Architecture}

The architecture of \emph{Focused Decoder} and its iterative refinement process are shown in Fig.~\ref{fig:foc_dec_arch}.
\emph{Focused Decoder} represents a lightweight Detection Transformer consisting solely of $N$ decoder blocks, omitting the Transformer's encoder completely. The Transformer's encoder has the task of modeling relations between elements of the input sequence via its self-attention module. In 3D, this task is highly complex due to long input sequences that arise from flattening high resolution 3D feature maps. Therefore, we argue that omitting the Transformer's encoder is a sound design choice when training on small-scale medical datasets. Following this hypothesis, \emph{Focused Decoder} forwards the input sequence given by the flattened feature map $x_{P2}$ directly to the decoder.

\emph{Focused Decoder} iteratively refines a set of object queries in $N$ stacked \emph{Focused Decoder} blocks (see Fig.~\ref{fig:foc_dec_arch} (right)). We assign an individual object query to each generated \emph{query anchor}, which results in $q_{\text{objects}} \in\mathbb{R}^{\#\text{classes} \times 27 \times d_{\text{hidden}}}$, containing class-specific object queries. By assigning object queries to unique spatial locations in the form of \emph{query anchors}, we demystify their associated learned positional embeddings, called query embeddings, and encourage diverse class-specific predictions, overcoming the issue of patient-to-patient variability. Besides the \emph{focused cross-attention module}, the general structure of the \emph{Focused Decoder} block remains similar to DETR's decoder. The main components of the \emph{Focused Decoder} block (see Fig.~\ref{fig:foc_dec_arch} (left)) are represented by a self-attention module followed by a \emph{focused cross-attention module} and a two-layer feedforward network (FFN), which utilizes a ReLU non-linearity in between its two linear layers. The FFN's in- and output dimensions correspond to the object queries' $d_{\text{hidden}}$. Its hidden dimension is represented by $d_{\text{FFN}}$. While the self-attention module aims to encode strong positional inter-dependencies among object queries, the \emph{focused cross-attention module} matches the input sequence to object queries and thus regulates the influence of individual feature map voxels for prediction via attention. Subsequently, the FFN facilitates richer feature representations. We employ additional residual connections and layer normalization operations to increase gradient flow.

The classification and bounding box regression head networks are attached to the last \emph{Focused Decoder} block and process the refined object queries (see Fig.~\ref{fig:foc_dec_arch} (left)).
Given that object queries are already preassigned to specific classes, the classification head predicts class-specific confidence scores, resulting in a binary classification task. % indicating the best suiting candidates per class for the current patient.
The classification head is given by a simple linear layer, whereas the bounding box regression head is represented by a more complex three-layer FFN, which predicts relative position and size offsets with regard to \emph{query anchors}. The hidden dimension of the bounding box regression head corresponds to the object queries' $d_{\text{hidden}}$. Eventually, we combine these relative offset predictions with their corresponding \emph{query anchors} and return the predicted bounding boxes together with the class-specific confidence scores, generating overall 27 candidate predictions per class.

Since the attention operation is known to be permutation invariant, valuable information about the spatial location of voxels would be dismissed. To tackle this issue, a positional encoding consisting of sine and cosine functions of different frequencies tries to encode information regarding the absolute position of voxels directly in the input sequence.

\subsubsection{Focused Cross-Attention}
At the core of \emph{Focused Decoder} lies the \emph{focused cross-attention module} forcing class-specific object queries to focus primarily on the structures of interest and their close proximity contained in RoIs of our anatomical region atlas. To this end, the \emph{focused cross-attention module} leverages an attention mask $M \in\mathbb{R}^{\# \text{queries} \times \# \text{voxels}}$ to restrict the object queries' FoV for cross-attention. This restriction of attention drastically simplifies the relation modeling task, as we solely have to learn dependencies between object queries and relevant voxels located in their respective RoIs. Equation~\eqref{eq:masked_attn} demonstrates the concept of masked attention, exploited in the \emph{focused cross-attention module}.
\begin{equation} \label{eq:masked_attn}
\text{MaskedAttn}(Q, K, V) = \text{softmax}(\frac{QK^{T}}{\sqrt{d_{\text{hidden}}}} + M)V
\end{equation}
Here, the key and value sequences $K$ and $V$ are derived from the input sequence, while the object queries contribute to the query sequence $Q$. Masking is done by adding the attention mask $M$ onto the raw attention weights $QK^{T} / \sqrt{d_{\text{hidden}}}$. The attention mask $M$ contains the value $- inf$ for voxels outside and the value $0$ for voxels inside the object queries' respective RoIs.
% In this context, valid attention weights result from dot-products between object queries and voxels located inside their respective RoIs. The remaining attention weights are consequently considered invalid.
By adding $- inf$, we nullify the importance of attention weights outside of RoIs for prediction. This is because attention weights corresponding to the value $- inf$ end up in extremely flat areas of the softmax function, which in turn kills gradient flow.
% This is because the dot-product between a query and voxel is smallest when both vectors are least similar.
% To generate $M$, we first create class-specific attention masks based on RoIs. Subsequently, we assign the class-specific attention masks to the corresponding class-specific object queries.

\subsubsection{Relative Offset Prediction} % we dont modify anchors!!!
\emph{Focused Decoder} leverages \emph{query anchors} to take advantage of the concept of relative offset prediction. To this end, we extend the bounding box regression head with a tanh nonlinearity, allowing object queries to modify their \emph{query anchors'} fixed center positions and sizes for prediction (see Fig.~\ref{fig:foc_dec_arch} (left)). However, these modifications are restricted. To counteract overlap and facilitate diverse predictions, center position offsets are restricted to a maximum value, resulting in distinct query-specific regions of possible final center positions covering the RoI completely (see Fig.~\ref{fig:anchor_gen} (right)). In contrast to position offset restrictions, which are derived from the sizes of RoIs, we restrict the allowed size offsets to the minimum and maximum bounding box sizes contained in our anatomical region atlas.

\subsubsection{Matching}
Since \emph{Focused Decoder} predicts 27 eligible candidates per class, matching boils down to finding the most suitable candidate among 27.
To train the head networks in a meaningful way, 
% we not only have to assign individual predictions to ground truth objects during training, but also generate ground truth confidence labels for the binary classification task. For this purpose, 
we generate dynamic confidence labels, solely relying on the normalized generalized intersection over union (GIoU) between all class-specific \emph{query anchors} and their corresponding ground truth objects. This results in dynamic confidence labels of 1 for predictions corresponding to \emph{query anchors} with the highest GIoU per class and dynamic confidence labels of 0 for predictions corresponding to \emph{query anchors} with the lowest GIoU per class. Based on these dynamic confidence labels, we match predictions with the highest dynamic confidence labels per class to the ground truth objects. We subsequently forward matched predictions and the dynamic confidence labels to the loss function.
% Since we have 27 dedicated predictions per class, the classification problem boils down to finding the best among 27, resulting in a binary classification of each object query. Unlike other Detection Transformers, which utilize a Hungarian matcher to uniquely assign individual predictions to the ground truth objects during training, \emph{Focused Decoder} has to assign one of the 27 predictions to the ground object for each class. For this purpose, we first estimate the generalized intersection over union (GIoU)  between all 27 \emph{query anchors} of each class and their corresponding ground truth box. Next, we generate dynamic objectness labels by normalizing the computed GIoU values for each class. This results in objectness labels 1 for \emph{query anchors} with the highest GIoU per class and objectness labels 0 for \emph{query anchors} with the lowest GIoU per class. The 27 predicted boxes per class inherit the class label from their corresponding \emph{query anchors}.
% Based on these dynamic objectness labels, we subsequently match predictions per class with the objectness labels of 1 to the ground truth object.
% Since \emph{Focused Decoder} pre-assigns object queries to RoIs and hence the anatomical structures present in the dataset, matching during inference is straight-forward. We simply select predictions with the highest objectness scores per class to represent the output.

\subsubsection{Loss Function}
Equation~\eqref{eq:foc_dec_loss} expresses \emph{Focused Decoder's} general loss function, where $N$ corresponds to the number of \emph{Focused Decoder} blocks. Following Deformable DETR, $\lambda_{\text{cls}}$, $\lambda_{\text{GIoU}}$, and $\lambda_{\ell1}$ correspond to 2, 2, and 5.
\begin{equation}\label{eq:foc_dec_loss}
\mathcal{L} = \sum_{n=1}^{N} (\lambda_{\text{cls}} \cdot \mathcal{L}_{\text{BCE}}^{n} + \lambda_{\text{GIoU}} \cdot \mathcal{L}_{\text{GIoU}}^{n} + \lambda_{\ell1} \cdot \mathcal{L}_{\ell1}^{n})
\end{equation}
We utilize a binary cross-entropy loss $\mathcal{L}_{\text{BCE}}$ between predicted confidence scores and dynamic confidence labels determined during matching to facilitate diverse predictions during inference. For bounding box regression, two loss functions, namely a scale-invariant GIoU loss $\mathcal{L}_{\text{GIoU}}$ and an $\ell1$ loss $\mathcal{L}_{\ell1}$, are combined. The bounding box regression loss functions solely consider matched predictions and thus only optimize predictions having dynamic confidence labels of 1. Following DETR, we additionally forward predictions generated based on outputs of earlier \emph{Focused Decoder} blocks (see Fig.~\ref{fig:foc_dec_arch} (right)) to the final loss function, resulting in an auxiliary loss for intermediate supervision.

\subsubsection{Inference}
During inference, the class-specific confidence scores predicted by the classification head indicate the most suitable candidates. Therefore, we simply select predictions with the highest confidence scores per class to represent the output.

%% file: sections/04_exp_res.tex
In this section, we determine \emph{Focused Decoder's} performance for 3D anatomical structure detection and compare it to two Detection Transformer baselines and a RetinaNet variant adopted from the state-of-the-art detection pipeline nnDetection.
% , adapted from the automated medical object detection pipeline nnDetection. 
Subsequently, we demonstrate the excellent explainability of \emph{Focused Decoder's} results and critically assess the importance of its design choices.

\subsection{Experimental Setup}
\begin{figure}[t]
\centerline{\includegraphics[width=\linewidth]{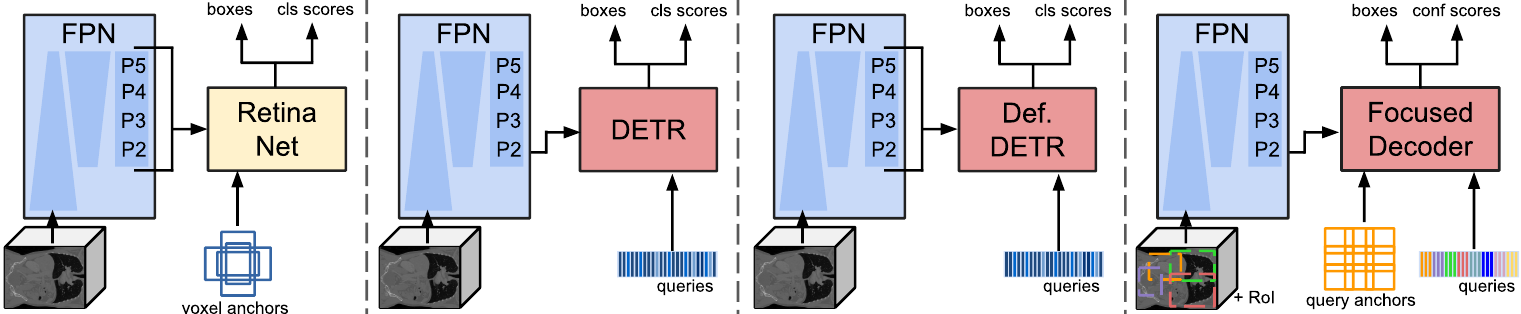}}
\vspace{-1.0em}
\caption{Overview of compared detectors. All featured architectures utilize the same FPN feature extraction backbone. Detection Transformers are depicted in red, while CNN-based traditional approaches are shown in yellow.}
\label{fig:method_overview}
\vspace{-1.5em}
\end{figure}

First, it should be mentioned that comparability of performances is of utmost importance and, due to the lack of 3D anatomical structure detection benchmarks, challenging to achieve.
Therefore, we adapt and directly integrate detectors featured in this work into the same detection and training pipeline for a fair and reproducible comparison.
%and validate their performances on two publicly available segmentation datasets.
For example, all featured detectors were trained until convergence on a single Quadro RTX 8000 GPU using the same AdamW optimizer, step learning rate scheduler, data augmentation techniques, and FPN feature extraction backbone.
In addition, we tried to keep the configurations of featured Detection Transformers as similar as possible. We use the same head networks, set $N$ to three, $d_\text{hidden}$ to 384, and $d_{\text{FFN}}$ to 1024. By doing so, we ensure maximal comparability of results. Precise information about training details, hyperparameters, and individual model configurations is available at \url{https://github.com/bwittmann/transoar}. To facilitate understanding of our experimental setup, Fig.~\ref{fig:method_overview} depicts an overview of all featured detectors.
% All models were trained on a single Quadro RTX 8000 GPU. We set the batch size to two and employed the AdamW optimizer with a weight decay of $10^{-4}$. After an appropriate amount of training epochs, the learning rate is dropped via a step learning rate scheduler that multiplies current learning rates by a factor of 0.1.\\
Next, we briefly introduce the Detection Transformer baselines, which we adapted for 3D anatomical structure detection, and our RetinaNet variant.
% Specifically, we state the configurations of our baseline Detection Transformers and the CNN-based state-of-the-art model. 
% In general, we tried to keep DETR's and Deformable DETR's configurations, if the memory constraint of the Quadro RTX 8000 GPU allowed it, as close as possible to \emph{Focused Decoder}'s configuration by employing, for example, the same head networks and the same hidden dimensions.

\subsubsection{DETR}\label{exp_detr}
DETR acts as our first Detection Transformer baseline. 
%Since DETR was initially developed to process 2D natural images, we adapt its architecture to process 3D voxel grids instead. However,
Besides some additional minor adjustments, the general structure and configuration of our DETR variant remain essentially unchanged.
%We experiment with a lightweight DETR variant. To this end, DETR's encoder-decoder architecture consists of three standard encoder and decoder blocks. 
Following the original DETR's configuration, we set the number of object queries, which are in contrast to \emph{Focused Decoder} not preassigned to specific classes, to 100. While the Hungarian algorithm is responsible for matching during training, we return the highest-scoring predictions per class during inference.

\subsubsection{Deformable DETR}
To evaluate \emph{Focused Decoder} properly, we introduce Deformable DETR as an additional Detection Transformer baseline. 
%For this purpose, we adjusted Deformable DETR to process 3D voxel grids instead of 2D natural images.
Deformable DETR generates its input sequence by flattening and combining the multi-level feature maps of the stages P2 to P5 into a combined input sequence and refines, similar to DETR, a set of 100 object queries.
%Following \emph{Focused Decoder}, we experiment with a lightweight Deformable DETR variant, which consists of three deformable decoder and three deformable encoder blocks, and refines a set of 100 object queries. 
Deformable DETR employs the same matching strategies during inference and training as DETR.

\subsubsection{RetinaNet}
% Since the Retina U-Net variant featured in the automated detection pipeline nnDetection has shown to deliver state-of-the-art results for numerous medical detection task, we adopt Retina U-Net from nnDetection with minor modifications and convert it to a RetinaNet variant by omitting Retina U-Net's segmentation proxy task
As a representative of CNN-based detectors, we adopt Retina U-Net from nnDetection with minor modifications. Since the automated detection pipeline nnDetection developed around Retina U-Net has been shown to deliver state-of-the-art results for numerous medical detection tasks, we extract nnDetection's generated hyperparameters and refine them via a brief additional hyperparameter search. In addition, we omit Retina U-Net's segmentation proxy task, converting it to a RetinaNet variant. This not only enforces comparability with other detectors featured in this work but also drastically reduces training times. We justify this decision as experiments\footnote{We provide experiments with the segmentation proxy task incorporated in all featured detectors in our GitHub repository. To activate the segmentation proxy task, set flag \texttt{use\_seg\_proxy\_loss} to \texttt{True} in the respective config file.} have shown that the segmentation proxy task leads to extremely minor and hence negligible performance benefits for 3D anatomical structure detection.
% The bounding box regression and classification head networks process the multi-level feature maps of the stages P2 to P5 in parallel.

\subsection{Datasets}
Due to the lack of 3D anatomical structure detection benchmarks, we conduct experiments on two CT datasets of well-defined FoV that were originally developed for the task of semantic segmentation, namely the VISCERAL anatomy benchmark~\citep{visceral} and the AMOS22 challenge~\citep{amos}. We transform their voxel-wise annotations into bounding boxes and class labels in the dataloader. Dataset-specific spatial sizes, approximated resolutions, and sizes of the respective training, validation, and test sets are along with the total number of subjects reported in Table~\ref{tab:datasets}.

\begin{table}[H]
\vspace{-0.5em}
\centering
\scriptsize
\caption{Final dataset properties after preprocessing.}
\label{tab:datasets}
\begin{tabular}{c|c c c c} 
\toprule
Dataset & Size & Resolution (in mm) & Train / Val / Test & \# of subjects\\
\midrule
VISCERAL & $160 \times 160 \times 256$ & $2.5\times 2.1\times 2.7$ & 120 / 20 / 12 & 152\\
AMOS22 & $256 \times 256 \times 128$ & $1.0\times 0.7\times 3.5$ & 117 / 20 / 18 & 155\\
\bottomrule
\end{tabular}
\end{table}

\subsubsection{VISCERAL Anatomy Benchmark}
The VISCERAL anatomy benchmark contains segmentations of 20 major anatomical structures. The CT images of the silver corpus subset are used as training data, while the gold corpus subset is split into two halves to create the validation and test sets.

\subsubsection{AMOS22 Challenge}
The multi-modality abdominal multi-organ segmentation challenge of 2022, short AMOS22, provides CT images with voxel-level annotations of 15 abdominal organs. We divide the CT images of the challenge's first stage into the training, validation, and test sets.

\subsubsection{Preprocessing}
We utilize the same preprocessing approach for both datasets. First, the raw CT images and labels are transformed to a uniform orientation represented by the RAS, short for 'right, anterior, superior', axis code. We subsequently crop CT images to foreground structures. By doing so, we additionally try to ensure datasets consisting of images with a similar FoV, which is necessary to determine meaningful class-specific RoIs. In the next step, the CT images and their labels are resized to a fixed spatial size. This increases fairness and further reduces the detection task's complexity by compensating for variations in patient body size. To compensate for incompletely labeled CT images, we discard CT images based on a specified label threshold, allowing for a sufficient amount of data while simultaneously ensuring fairly clean datasets. Additionally, we completely omit partially labeled CT images in the test sets to provide a valid performance estimate.

\subsection{Metric}
The metric commonly used to evaluate object detection algorithms is the mean average precision (mAP).
The mAP metric, described in~\eqref{eq:map}, is defined as the mean of a selected subset of average precision (AP) values. We report $\text{mAP}_{\text{coco}}$, which is calculated based on AP values evaluated at IoU thresholds $\mathbb{T}$ = \{0.5, 0.55, ..., 0.95\}.
\begin{equation}\label{eq:map}
\begin{split}
\text{mAP}_{\text{coco}} = \frac{1}{|\mathbb{T}|} \sum_{t \in \mathbb{T}} \text{AP}_{t}\\
\end{split}
\end{equation}
To determine detection performance related to structure size, we introduce $\text{mAP}_{\text{coco}}^{\text{S}}$, $\text{mAP}_{\text{coco}}^{\text{M}}$, and $\text{mAP}_{\text{coco}}^{\text{L}}$. To this end, we categorize classes based on the volume occupancy of their median bounding boxes into the subsets S (small), M (medium), and L (large), which rely on the volume occupancy ranges of [0.0\%, 0.5\%), [0.5\%, 5.0\%), and [5.0\%, 100\%], respectively. Subsequently, we reevaluate $\text{mAP}_{\text{coco}}$ restricted to solely classes in the individual subsets. The dataset-specific subsets determined based on the above reported volume occupancy ranges are shown in Table~\ref{tab:size_specific_map}. It should be mentioned that although both datasets have anatomical structures in common, they might be assigned to different subsets due to varying FoVs. The difference in FoV across both datasets can be observed in Fig.~\ref{fig:qual_res} (left).

\begin{table}[ht]
\centering
\scriptsize
\caption{Size-specific subsets.}
\label{tab:size_specific_map}
\renewcommand{\arraystretch}{1.5}
\begin{tabular}{cc|l} 
\toprule
% Dataset & Subset & Classes\\
% \midrule
\multirow{3}{*}{\rotatebox[origin=c]{90}{VISCERAL}}
& S & \makecell[l]{pancreas, gall bladder, urinary bladder, trachea, thyroid gland,\\ first lumbar vertebra, adrenal glands}\\
& M & \makecell[l]{spleen, aorta, sternum, kidneys, psoas major, rectus abdominis}\\
& L & liver, lungs\\
\midrule
\multirow{3}{*}{\rotatebox[origin=c]{90}{AMOS}}
& S & esophagus, adrenal glands, prostate/uterus\\
& M & \makecell[l]{spleen, kidneys, gall bladder, aorta, postcava, pancreas, duodenum,\\ urinary bladder}\\
& L & liver, stomach\\
\bottomrule
\end{tabular}
\end{table}

\begin{table}[t]
\centering
\vspace{-1.0em}
\scriptsize
\caption{Quantitative results estimated on the VISCERAL anatomy benchmark and the AMOS22 challenge. We compare \emph{Focused Decoder} to the baseline architectures DETR, Deformable DETR, and RetinaNet. 
\emph{Focused Decoder} performs best on large structures, while RetinaNet dominates on small structures.
}
\label{tab:quantitative_results}
\begin{tabular}{p{50pt}|l|c|c|c|c|c} 
\toprule
Dataset & Model &$\#\text{params}\downarrow$ &$\text{mAP}_{\text{coco}}\uparrow$ &$\text{mAP}_{\text{coco}}^{\text{S}}\uparrow$ &$\text{mAP}_{\text{coco}}^{\text{M}}\uparrow$ &$\text{mAP}_{\text{coco}}^{\text{L}}\uparrow$\\
\midrule
\multirow{4}{*}{VISCERAL}
& RetinaNet & 52.8M & \textbf{41.37}* & \textbf{20.99}* & \textbf{47.02} & 78.74\\
% \cline{2-9}
\cmidrule{2-7}
& DETR & 46.5M & 27.35 & 11.32 & 28.10 & 67.88\\
& Def. DETR & 54.8M & 33.06 & 14.52 & 35.09 & 76.39\\
& Focused Decoder & \textbf{41.8M} & 39.22 & 18.33 & 43.59 & \textbf{81.70}*\\
\midrule
\multirow{4}{*}{AMOS22}
& RetinaNet & 52.1M & \textbf{30.38} & \textbf{16.92}* & 32.37 & 44.38\\
\cmidrule{2-7}
& DETR & 43.7M & 16.47 & 4.24 & 15.68 & 32.96\\
& Def. DETR & 53.4M & 26.59 & 8.93 & 30.32 & 43.08\\
& Focused Decoder & \textbf{42.6M} & 29.83 & 10.30 & \textbf{34.01} & \textbf{47.97}*\\
\bottomrule
\multicolumn{7}{l}{* denotes statistically significant difference between Focused Decoder and RetinaNet; p-values $<0.05$.}
\end{tabular}
\vspace{-1.0em}
\end{table}

\subsection{Quantitative Results}
Table~\ref{tab:quantitative_results} lists quantitative results of all detectors featured in this work. Essentially, our findings remain consistent across both datasets. \emph{Focused Decoder}, possessing the least amount of trainable parameters, significantly outperforms the Detection Transformer baselines, namely DETR and Deformable DETR, and performs close to the state-of-the-art detector RetinaNet.

This demonstrates that a naive adaption of 2D Detection Transformers to the medical domain is definitely not sufficient to overcome the problem of data scarcity. However, \emph{Focused Decoder's} strong detection performances reveal that reducing the complexity of the relation modeling and thus the 3D detection task alleviates the need for large-scale annotated datasets, converting Detection Transformers into competitive and lightweight (approximately $20 \%$ fewer parameters compared to RetinaNet) 3D medical detection algorithms. \emph{Focused Decoder} not only performs close to or on par with its CNN-based counterpart but also provides the most accurate predictions for larger structures. On the other hand, RetinaNet outperforms \emph{Focused Decoder} on smaller structures, overcoming the Detection Transformer's well-known shortcoming of poor small object detection performance, which is subject to open research.

Even though the overall detection performance on the VISCERAL anatomy benchmark turns out to be stronger, \emph{Focused Decoder} manages to narrow the gap to RetinaNet on the AMOS22 challenge. The generally stronger performances on the VISCERAL anatomy benchmark can be attributed to the absence of symmetrical and easy to detect structures such as the lungs or the rectus abdominis in the AMOS22 challenge. The narrowed performance gap, however, primarily results from the difference in FoV between both datasets. This is because the reduced FoV of the AMOS22 challenge (see  Fig.~\ref{fig:qual_res} (left)) leads to an artificial increase in structure size, which is most beneficial for \emph{Focused Decoder}.

% Even though the overall detection performance on the VISCERAL anatomy benchmark turns out to be stronger compared to the AMOS22 challenge, \emph{Focused Decoder} manages to narrow the gap to RetinaNet on the AMOS22 challenge. This observation can be traced back to two properties of the AMOS22 challenge. First, the absence of large and easy to detect structures such as the lungs or the rectus abdominis, lowering general detection performance; and second, the CT input images' reduced FoV. The reduction in FoV can be observed in Fig.~\ref{fig:qual_res} (left) and leads to an artificial increase in structure size, which is most beneficial for \emph{Focused Decoder}.\\
% Interestingly, Deformable DETR outperforms DETR by a noteworthy margin. This observation can be traced back to the deformable attention mechanism, which represents a sparse operation as deformable attention only models relations between a query and a small set of sampled key points. Nevertheless, quantitative results prove that the concept of focused attention remains most applicable for medical applications compared to DETR's complete attention and Deformable DETR's deformable attention.
To estimate the statistical significance of the performance differences between \emph{Focused Decoder} and RetinaNet, we determined p-values using the Wilcoxon signed-rank test based on bootstrapped subsets. 
% of the combined validation and test sets. 
We indicate statistically significant performance differences in Table~\ref{tab:quantitative_results} (see footnote).
% We indicate \emph{Focused Decoder}'s quantitative results corresponding to comparisons of high statistical significance in Table~\ref{tab:quantitative_results}.
% Although a p-value of 1.34\% confirms RetinaNet's dominance on the VISCERAL anatomy benchmark, we argue that a p-value of 34.76\% indicates that \emph{Focused Decoder} potentially outperforms RetinaNet on the AMOS22 challenge.

% \begin{figure}[h]
% \centerline{\includegraphics[width=0.5\columnwidth]{imgs/attention_weights4.pdf}}
% \caption{Focused Decoder's cross-attention weights (red) corresponding to selected anatomical structures (black) of the VISCERAL anatomy benchmark. Upon closer inspection, one can identify the actual class-specific RoIs, restricting the cross-attention's FoV. The class-specific RoIs are especially visible for small and hence hard to detect structures, as the model's uncertainty is reflected in less precise cross-attention weights.}
% \label{fig:attn_weights}
% \end{figure}

% \begin{figure}[h]
% \centerline{\includegraphics[width=0.5\columnwidth]{imgs/attention_weights_comp6.pdf}}
% \caption{Comparison of explainability of results. We compare the explainability of Focused Decoder's, DETR's, and RetinaNet's predictions corresponding to the urinary bladder and the left adrenal gland.}
% \label{fig:attn_weights_comp}
% \end{figure}

\subsection{Explainability of Focused Decoder's Predictions}
Explainability of results is of exceptional importance in the context of medical image analysis. While identifying reasons behind predictions produced by CNN-based detectors is often cumbersome and requires additional engineering susceptible to errors, \emph{Focused Decoder's} attention weights facilitate the accessibility of explainable results. Therefore, we particularly probe into intra-anatomical and inter-anatomical explainability by analyzing cross- and self-attention weights.

\begin{figure}[h]
\centerline{\includegraphics[width=1.0\columnwidth]{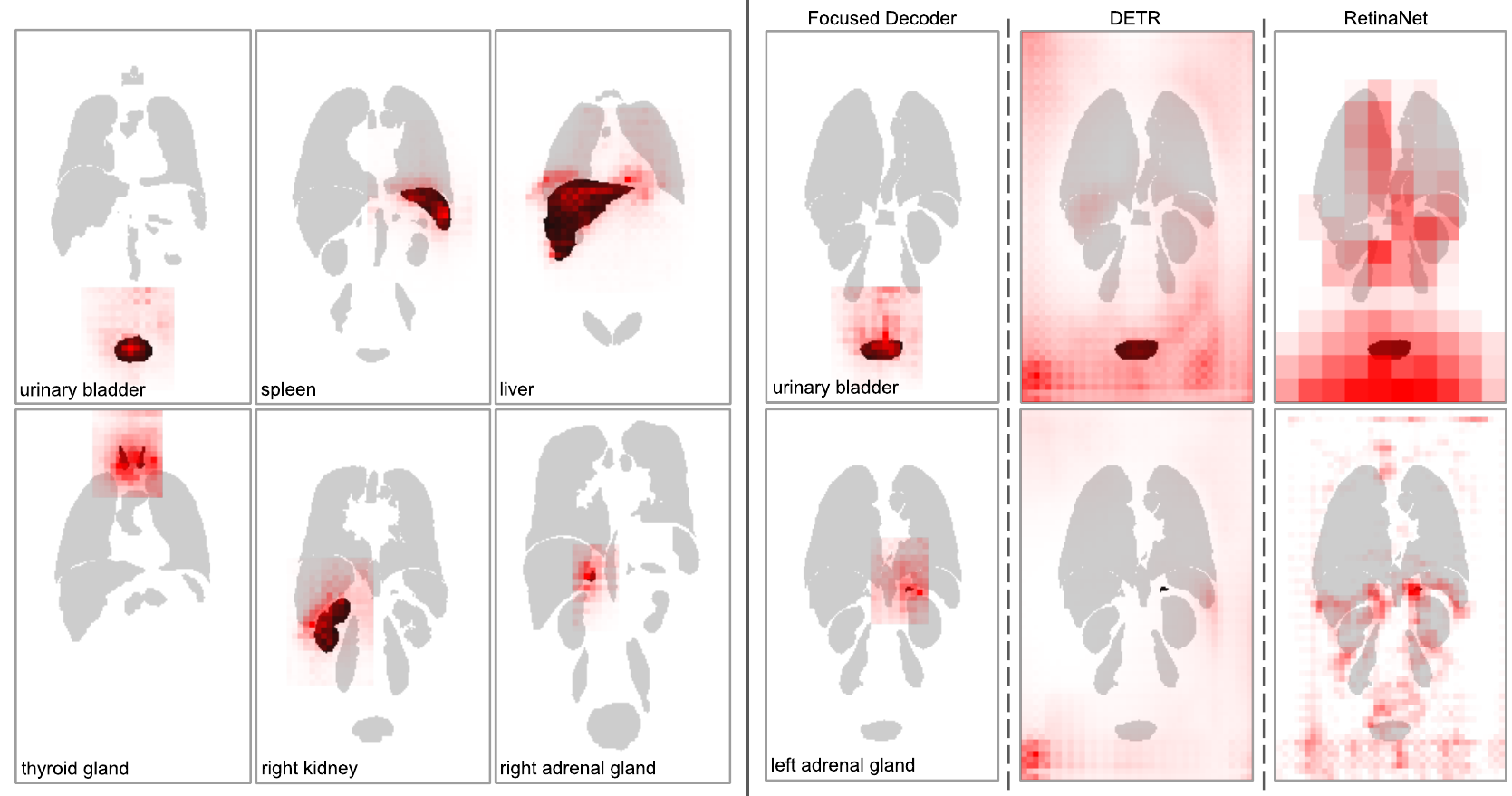}}
\vspace{-1.0em}
\caption{
(\textbf{Left}): \emph{Focused Decoder's} cross-attention weights (red) corresponding to six selected anatomical structures (black) of the VISCERAL anatomy benchmark. Upon closer inspection, one can identify the actual class-specific RoIs, restricting the cross-attention's FoV. The class-specific RoIs are especially visible for small and hence hard to detect structures, as the model's uncertainty is reflected in less precise cross-attention weights.
(\textbf{Right}): Comparison of explainability of results. We compare the explainability of \emph{Focused Decoder's}, DETR's, and RetinaNet's predictions corresponding to the urinary bladder (first row) and the left adrenal gland (second row).
}
\label{fig:attn_weights_combined}
\vspace{-1.0em}
\end{figure}

\subsubsection{Intra-Anatomical Explainability}
\emph{Focused Decoder's} cross-attention weights between the input sequence and object queries corresponding to selected anatomical structures are visualized in Fig.~\ref{fig:attn_weights_combined} (left). Since cross-attention weights indicate the importance of voxels for prediction, one can conclude that \emph{Focused Decoder's} predictions primarily rely on structures of interest and the context in their proximity, even for extremely small objects such as the adrenal glands.
% Furthermore, it should be emphasized that \emph{Focused Decoder} consistently generates precise, sparse cross-attention weights even for extremely small objects such as the adrenal glands.

To highlight \emph{Focused Decoder's} intra-anatomical explainability of results, we additionally compare its cross-attention weights to cross-attention weights generated by DETR and gradient-weighted class activation maps corresponding to RetinaNet's predictions in Fig.~\ref{fig:attn_weights_combined} (right). These gradient-weighted activation maps were generated using the established Grad-CAM~\citep{gradcam} algorithm, backpropagating to the FPN's output feature maps. \emph{Focused Decoder's} cross-attention weights provide the best explainability of results. In contrast, DETR's cross-attention weights and RetinaNet's class activations are scattered over large parts of the CT image, complicating explainability and possibly indicating memorization of dataset-specific covariates.
% The superior performance of \emph{Focused Decoder} compared to the Detection Transformer baselines is also reflected in cross-attention weights. In general, cross-attention weights should be sparse and focus mainly on the object of interest and the context in its proximity. This is not only essential for explainability of results, but also for detection performance, as cross-attention weights regulate the influence of individual feature map voxels for prediction. Even though focusing on the legs and the liver should give a rough estimate of the urinary bladder's location (see DETR's cross attention-weights), precise localization requires a strong focus on the organ itself (see \emph{Focused Decoder}'s cross-attention weights).

\subsubsection{Inter-Anatomical Explainability}
\begin{figure}[H]
\centerline{\includegraphics[width=0.55\columnwidth]{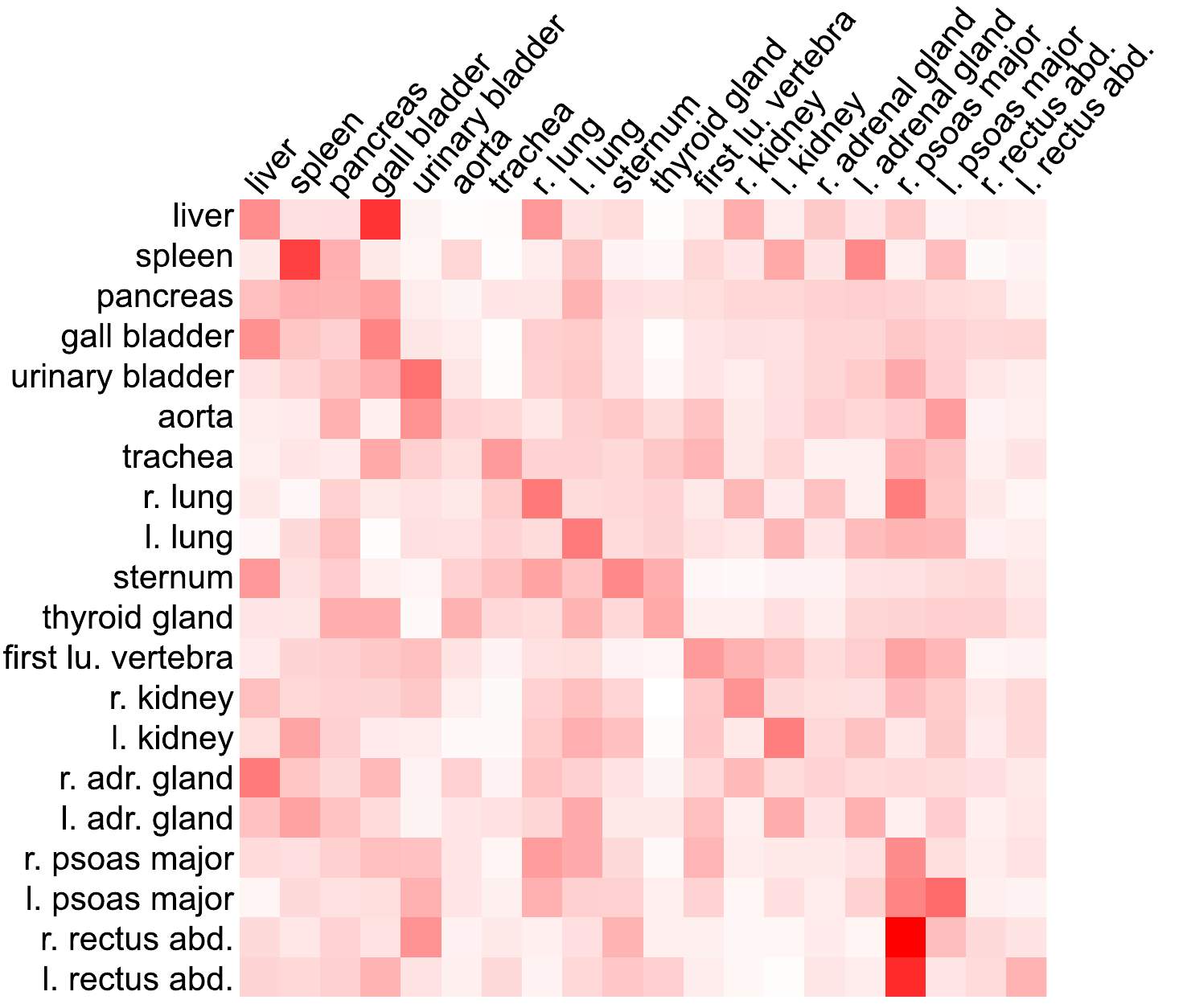}}
\vspace{-1.0em}
\caption{\emph{Focused Decoder’s} self-attention weights between class-specific object queries corresponding to structures of the VISCERAL anatomy benchmark. We averaged attention weights of object queries corresponding to the same class to provide a concise
visualization.}
\vspace{-1.0em}
\label{fig:self_attn_weights}
\end{figure}

While cross-attention weights indicate the importance of voxels for prediction, self-attention weights reveal the inter-dependencies among the class-specific object queries, providing inter-anatomical explainability. We visualize self-attention weights in Fig.~\ref{fig:self_attn_weights} to prove the hypothesis that \emph{Focused Decoder} captures meaningful inter-dependencies among anatomical structures.

% One can conclude that \emph{Focused Decoder} succeeds in capturing meaningful inter-dependencies.
We observe the trend of high inter-dependencies among neighboring structures. Object queries corresponding to the liver (first row), for example, attend primarily to structures in the liver's proximity, such as the gall bladder, the right lung, the right kidney, and the right psoas major, whereas object queries corresponding to the spleen (second row) attend primarily to the pancreas, the left lung, the left kidney, and the left adrenal gland.

\subsection{Qualitative Results}
\begin{figure}[t]
\centerline{\includegraphics[width=\linewidth]{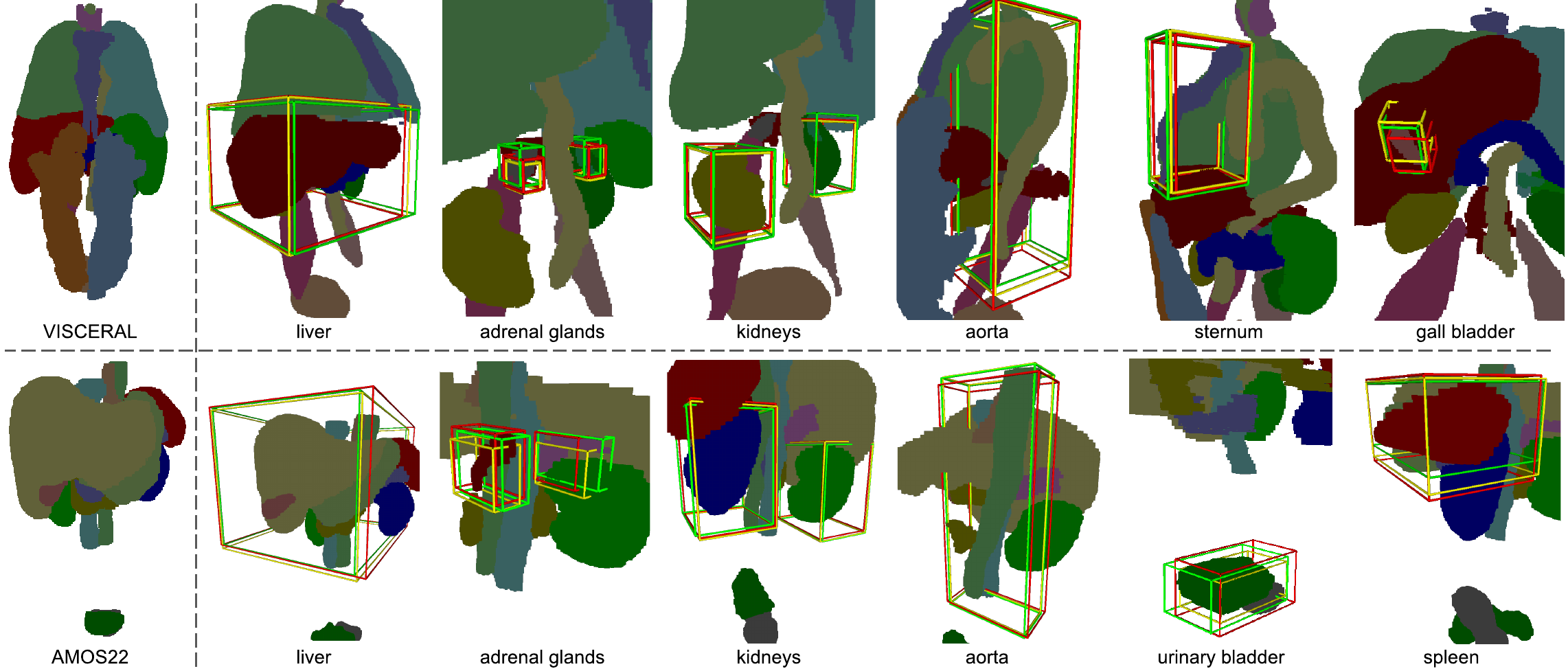}}
\caption{Qualitative results. We compare RetinaNet's predictions (red) and \emph{Focused Decoder's} predictions (yellow) to the ground truth (green) on the VISCERAL anatomy benchmark (first row) and the AMOS22 challenge (second row).}
\label{fig:qual_res}
\end{figure}

Qualitative results in the form of predicted bounding boxes of the two best-performing architectures, RetinaNet and \emph{Focused Decoder}, are compared to the ground truth in Fig.~\ref{fig:qual_res}. One can observe that \emph{Focused Decoder} exhibits exceptional detection performance on larger structures such as the liver or the aorta, while RetinaNet remains superior in detecting small structures such as the adrenal glands. 
%Moreover, it should be pointed out that the estimated bounding boxes are precise and can therefore, including a certain margin, be utilized to exploit the two-stage segmentation approach motivated in Section~\ref{sec:introduction}.

\subsection{Ablation Studies}
In this subsection, we investigate the importance of \emph{Focused Decoder's} design choices by conducting detailed ablations on the validation set of the VISCERAL anatomy benchmark. Table~\ref{tab:ablations_des} displays $\textrm{mAP}_{\textrm{coco}}$ values produced by \emph{Focused Decoder's} base configuration (first row) and three additional configurations (second to fourth row), omitting different design choices.

\begin{table}[H]
\centering
\scriptsize
\caption{Ablation on \emph{Focused Decoder's} main design choices.}
\label{tab:ablations_des}
\begin{tabular}{c c c|c c} 
\toprule
Restriction & Anchors & Queries per class & $\text{mAP}_{\text{coco}}\uparrow$ &$\Delta$ \\
\midrule
\checkmark & \checkmark & 27 & \textbf{37.78} & $-$\\
\checkmark &  & 27 & 36.63 & -1.15\\
\checkmark &  & 1 & 35.08 & -2.70\\
 & \checkmark & 27 & 32.03 & -5.75\\
\bottomrule
\end{tabular}
\end{table}

To evaluate the impact of our proposed \emph{query anchors}, we completely omit them in the second configuration, resulting in an $\textrm{mAP}_{\textrm{coco}}$ decrease of 1.15. This proves the importance of demystifying object queries and their associated query embeddings by assigning them to precise spatial locations and thus generating diverse class-specific predictions, overcoming the issue of patient-to-patient variability.
The configuration presented in the third row additionally reduces the amount of object queries per class from 27 to one. Based on the $\textrm{mAP}_{\textrm{coco}}$ reduction of 2.70, we argue that having only one object query per class fails to cover all class-specific object variations and hence leads to performance decreases.
Next, we deactivate the \emph{focused cross-attention module's} restriction to RoIs in the fourth row. As expected, detection performance drastically diminishes, which is reflected in an $\textrm{mAP}_{\textrm{coco}}$ delta of -5.75. The result of this ablation study repeatedly demonstrates the necessity of simplifying the relation modeling task to achieve competitive detection performances.

Since the FPN outputs a set of refined multi-level feature maps $\{x_{Pl} \}_{l=2}^{5}$, we experiment with different input sequences represented by flattened feature maps of different resolutions and report our findings in Table~\ref{tab:ablations_fmap}.

\begin{table}[H]
\centering
\scriptsize
\caption{Ablation on input feature maps.}
\label{tab:ablations_fmap}
\begin{tabular}{c c|c c} 
\toprule
Stage & Feature map resolution & $\text{mAP}_{\text{coco}}\uparrow$ & $\Delta$\\
\midrule
P2 & $40\times40\times64$ &\textbf{37.78} & $-$\\
P3 & $20\times20\times32$ &36.97 & -0.81\\
P4 & $10\times10\times16$ &33.72 & -4.07\\
P5 & $5\times5\times8$ &29.96 & -7.82\\
\bottomrule
\end{tabular}
\end{table}

One can observe that feature maps of higher resolution are clearly beneficial for detection performance, as they contain more fine-grained details, which in turn leads to more precise bounding box estimations. However, the performance improvement saturates as the feature map resolutions increase.

Finally, we prove that omitting the Transformer's encoder leads to more precise detections when the amount of available training data is strictly limited. To this end, we experiment with DETR and Deformable DETR configurations omitting the encoder and report the results in Table~\ref{tab:ablations_enc}.

\begin{table}[H]
\centering
\scriptsize
\caption{Impact of the Transformer's encoder.}
\label{tab:ablations_enc}
\begin{tabular}{c c|c c c c} 
\toprule
Model & Encoder & $\text{mAP}_{\text{coco}}\uparrow$ & $\text{AP}_{50}\uparrow$ &$\text{AP}_{75}\uparrow$\\
\midrule
DETR &  & \textbf{25.94} & \textbf{61.49} & \textbf{17.46}\\
DETR & \checkmark & 23.32 & 59.11 & 14.42\\
\midrule
Def DETR &  & \textbf{30.95} & \textbf{67.07} & \textbf{24.35} \\
Def DETR & \checkmark & 29.26 & 65.31 & 19.44\\
\bottomrule
\end{tabular}
\end{table}

Evaluation of $\textrm{mAP}_{\textrm{coco}}$ values presented in Table~\ref{tab:ablations_enc} leads to the conclusion that the Transformer's encoder is disruptive to detection performance. This supports the hypothesis that the encoder's intricate relation modeling task would require an extreme amount of training epochs and hence a large amount of annotated data to capture meaningful dependencies among the input sequence elements.

%% file: sections/05_limitations.tex
We would like to particularly highlight the limitations of \emph{Focused Decoder's} design, which can be primarily attributed to the assumption of well-defined FoVs. Although this assumption is reasonable to make from a clinical standpoint, it also poses a significant challenge to \emph{Focused Decoder's} robustness with regard to CT images of varying FoVs. This is because drastic shifts in FoV may lead to anatomical structures being located partially outside of RoIs, which in turn may hinder object queries to encode meaningful information via our proposed \emph{focused cross-attention module}. However, it should be emphasized that the FoV of CT images could always be adjusted to fit the FoVs of our datasets by adopting our preprocessing step or based on registration via a few anatomical landmarks.

It is also worth mentioning that \emph{Focused Decoder} predicts by design for each anatomical structure contained in our anatomical region atlases exactly one final bounding box. Therefore, the complete absence of anatomical structures, for example after radical nephrectomy, would remain unnoticed.

% , exceeding the boundaries of RoIs contained in the anatomical region atlas.

% This is because CT images pushing the limitations of the assumption of well-defined FoVs
% , which can be traced back to the assumption of well-defined FoVs. Although this is a reasonable assumption to make, it is often hard to ensure. Therefore,  
% , which is resembled by the robustness of the RoI. Even though we tried to ensure robust RoIs, a  assuming well-defined FoVs is a reasonable assumption from a clinical standpoint, there might be cases pushing these limitations.

%% file: sections/06_outlook.tex
This work lays the foundations of 3D medial Detection Transformers by introducing \emph{Focused Decoder}, a lightweight alternative to CNN-based detectors, which not only exhibits exceptional and highly intuitive explainability of results but also demonstrates the best detection performances on large structures. \emph{Focused Decoder's} impressive performances on large structures already conclusively indicate the immense potential of Detection Transformers for medical applications.

Based on results from our experiments, we recommend the use of \emph{Focused Decoder} for anatomical structure detection tasks when the priority lies on the analysis of large structures or the explainability of results. However, we strongly believe that with increasing sizes of annotated medical datasets, Detection Transformers will eventually outperform CNN-based architectures on medical detection tasks in all metrics, resulting in a complete shift from CNN- to Transformer-based architectures.

We encourage future work to further explore \emph{Focused Decoder's} parameters, overcome its limitations, and address its inferior performance on small structures by investigating the influence of the fixed FoV of CT images, which results in drastic size differences between anatomical structures (the right lung, for example, occupies magnitudes more space compared to the urinary bladder in CT images of fixed FoV). Future work should, therefore, also focus on exploring approaches operating on uniformly sized RoIs and hence dynamic spatial resolutions. This would in turn lead to uniform structure sizes, possibly allowing \emph{Focused Decoder} to overcome the issue of relative scale between anatomical structures.